\documentclass[conference]{IEEEtran}
\IEEEoverridecommandlockouts

\usepackage{multirow}

\usepackage{booktabs}

\usepackage{adjustbox}
\usepackage{makecell}

\usepackage{algorithm}
\usepackage{algpseudocode}

\usepackage{xparse}

\usepackage{pgffor}
\usepackage{etoolbox}

\usepackage{lipsum}

\usepackage{float}

\usepackage{enumitem}

\usepackage{colortbl}
\usepackage{graphicx}
\usepackage{caption}
\usepackage{subcaption}
\usepackage{xspace}

\usepackage{arydshln}

\definecolor{my_red}{RGB}{243, 181, 179}
\definecolor{my_orange}{RGB}{248, 218, 182}
\definecolor{my_yellow}{RGB}{255, 254, 185}
\newcommand{\best}{\cellcolor{my_red}}
\newcommand{\sbest}{\cellcolor{my_orange}}
\newcommand{\tbest}{\cellcolor{my_yellow}}

\usepackage{amsthm}

\newtheorem{proposition}{Proposition}

\usepackage{tocloft}
\usepackage{titlesec}

\usepackage[hidelinks]{hyperref}

\usepackage{cite}
\usepackage{amsmath,amssymb,amsfonts}
\usepackage{graphicx}
\usepackage{textcomp}
\usepackage{xcolor}
\def\BibTeX{{\rm B\kern-.05em{\sc i\kern-.025em b}\kern-.08em
    T\kern-.1667em\lower.7ex\hbox{E}\kern-.125emX}}
\begin{document}

\title{Dynamic Importance in Diffusion U-Net for Enhanced Image Synthesis}

\author{
    \href{mailto:hytidel333@gmail.com}{Xi Wang} \qquad \href{mailto:autumngoosehe@gmail.com}{Ziqi He} \qquad \href{mailto:zhouyangvcc@gmail.com}{Yang Zhou$^\dag$}
    \thanks{$^\dag$ Corresponding author. } \\
	CSSE, Shenzhen University, Shenzhen, China \\
}

\makeatletter
\let\@oldmaketitle\@maketitle
\renewcommand{\@maketitle}{\@oldmaketitle
    \vspace{2\baselineskip}
    \setcounter{figure}{0}
        \vspace{-2em}
    \includegraphics[width=1.0\linewidth]{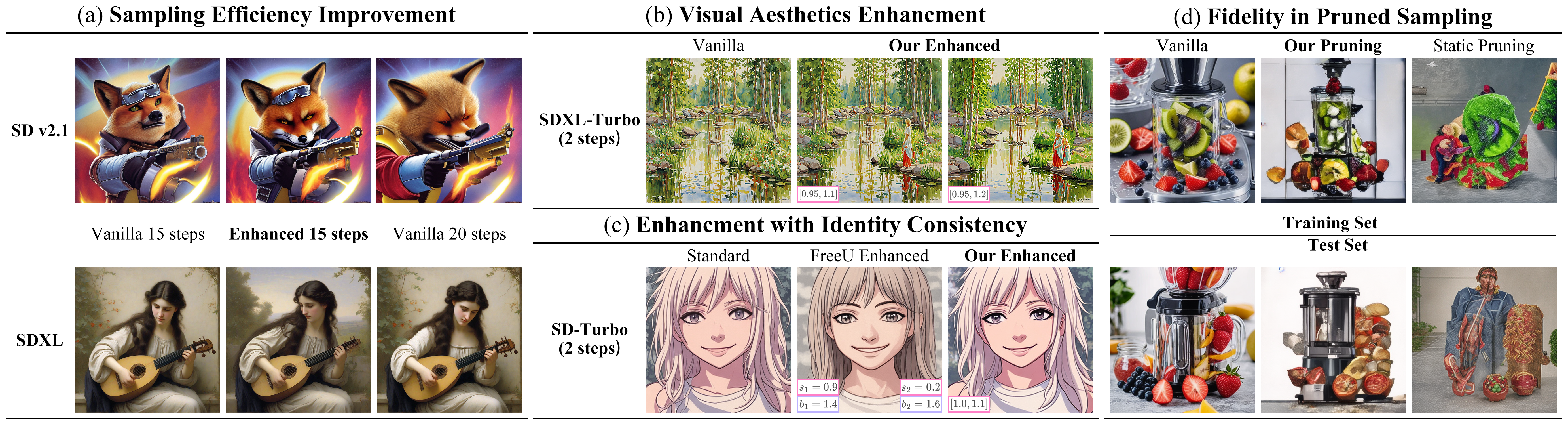}
    \captionof{figure}{
        Our approach 
        enhances the U-Net capability in the following tasks without additional training or fine-tuning:
        (a) improving sampling efficiency; 
        (b) \& (c) enhancing the visual aesthetics of samples with identity consistency; 
        and (d) achieving better fidelity in pruned sampling. 
        Images are evaluated at 512$\times$512/1024$\times$1024 px with the SD/SDXL model. 
    }
    \label{fig-head_figure}

    \vspace{-2\baselineskip}
\bigskip}

\maketitle

\begin{abstract}

Traditional diffusion models typically employ a U-Net architecture.
Previous studies have unveiled the roles of attention blocks in the U-Net. 
However, they overlook the dynamic evolution of their importance during the inference process, which hinders their further exploitation to improve image applications. 
In this study, 
we first theoretically proved that, re-weighting the outputs of the Transformer blocks within the U-Net is a ``free lunch'' for improving the signal-to-noise ratio during the sampling process. 
Next, we proposed \emph{Importance Probe} to uncover and quantify the dynamic shifts in importance of the Transformer blocks throughout the denoising process.
Finally, we design an adaptive importance-based re-weighting schedule tailored to specific image generation and editing tasks. 
Experimental results demonstrate that, our approach significantly improves the efficiency of the inference process, and enhances the aesthetic quality of the samples with identity consistency.  
Our method can be seamlessly integrated into any U-Net-based architecture. 
Code: \href{https://github.com/Hytidel/UNetReweighting}{https://github.com/Hytidel/UNetReweighting}

\end{abstract}

\begin{IEEEkeywords}

diffusion model, image synthesis, image editing

\end{IEEEkeywords}

\section{Introduction}

\begin{figure}[t]
    \centering
    \includegraphics[width=0.98\linewidth]{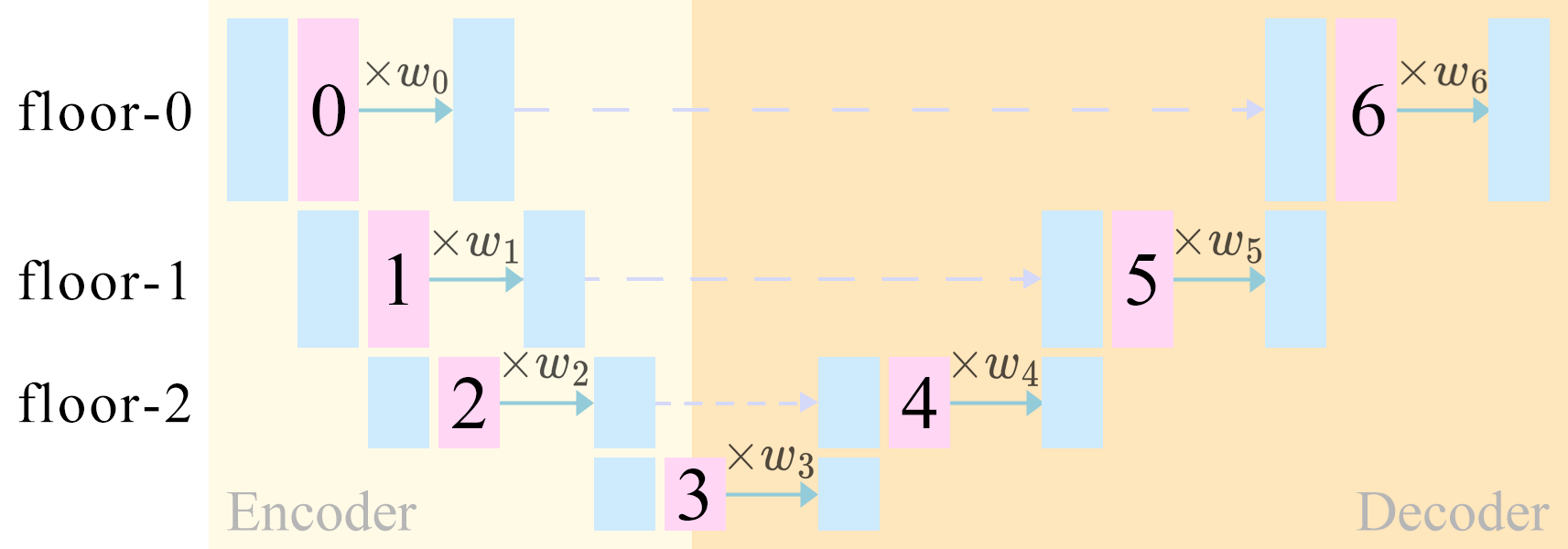}
    \caption{
        Illustration of how the outputs of Transformer blocks are scaled before being passed to subsequent ResNet blocks.
    }
    \label{fig-weighted_u_net}
    \vspace{-10pt}
\end{figure}

Diffusion Models (DMs)~\cite{ho2020denoisingdiffusionprobabilisticmodels, song2022denoisingdiffusionimplicitmodels} have emerged as exceptional performers in image generation. 
At the core of Stable Diffusion (SD)~\cite{rombach2022highresolutionimagesynthesislatent, sauer2024fasthighresolutionimagesynthesis} models, U-Nets play a pivotal role in predicting residual noise, which is typically structured symmetrically with a hierarchical architecture for multi-scale feature encoding and decoding (see Fig.~\ref{fig-weighted_u_net}).

Previous studies have revealed the roles of the attention blocks in the U-Net. 
It can be empirically summarized that, high-resolution blocks primarily focus on detail extraction, while mid-low-resolution blocks correspond to layout structuring and semantic understanding~\cite{pnpDiffusion2022}. 
Subsequent works on plug-and-play attention features show that, the U-Net also attends to features of different gratuities at variant denoising time steps~\cite{cao2023masactrl, zhou2024generating, zhou2025attentiondistillationunifiedapproach}.
Recently, FreeU~\cite{si2023freeufreelunchdiffusion} attempted to analyze the functionality of the attention blocks, showing that the backbone features and the skip connections of the U-Net contribute to information of different frequencies. Based on this finding, a re-weighting scheme is proposed to enhance the generation quality.
However, they overlook the dynamic shifts in block roles during the denoising process, which hinders their further exploitation.

In this paper, we propose \emph{Importance Probe} (IP), monitoring and quantifying the dynamic importance shifts of each Transformer block throughout the denoising process for the first time. 
Specifically, we first assign a non-negative weight to each U-Net Transformer block, and then dynamically adjust a weight threshold during denoising to probe their importance. 
We design a randomized heuristic search strategy to optimize the weight allocation by comparing the noise prediction errors between a student and a teacher U-Net, thus determining the \emph{importance rank} of each block.

Based on the importance ranking, we can re-weight the output of each Transformer block by a scaling factor before passing it to the subsequent block (see Fig.~\ref{fig-weighted_u_net}). We theoretically prove that our new re-weighting strategy enhances the signal-to-noise ratio (SNR) during the sampling process, which improves both the inference efficiency and the sample aesthetics. 
Note that the time-variant weights are selected based on the \emph{importance scores} derived from multiple runs of IP and aggregated using a voting mechanism. Therefore, our approach simultaneously accounts for the functional and importance variations of attention blocks. 

In experiments, we first validate the dynamic shifts in importance across blocks during the denoising process, as well as significant divergences in importance levels between symmetrically positioned blocks~(see Sec.~\ref{section-experiment_subsection-importance_ranking}). 
As no prior work has discussed block-level importance shifts in U-Nets, we verify our derived importance ranking through dynamic attention pruning (see Fig.~\ref{fig-head_figure} (d)).
Next, we apply our adaptive importance-based re-weighting schedule to text-to-image generation tasks. 
Specifically, for each prompt, we conduct several runs of IP and calculate the importance score for each Transformer block at every inference step. 
At each step, we assign weights slightly above $1.0$ to the dominant blocks, and weights slightly below $1.0$ to the less important blocks. 
Results demonstrated the effectiveness of our approach in reducing the number of inference steps while enhancing the visual aesthetic of samples with identity consistency (see Fig.~\ref{fig-head_figure} (a), (b) and (c)). 
Our approach can be seamlessly integrated into any U-Net-based DMs, showing the potential of incorporating dynamic mechanisms to improve the performance of DMs across various applications. 

\begin{table*}[h]
    \caption{
        Comparison between different models and inference steps. 
        Cells with a red/orange/yellow background indicate the best/second-best/third-best performance, respectively.  
        Cells where the weighted performance is worse than the vanilla schedule are marked with a downward arrow $\downarrow$. 
        Blocks within the SD/SDXL family are numbered from 0/1 for symmetry. 
    }
    \centering
    \begin{tabular}{c | c c c : c c c | c c c : c c c}
        \hline
        \multirow{2}{*}{\textbf{Weighting}} &
        \multicolumn{3}{c:}{\textbf{SD-Turbo}} & 
        \multicolumn{3}{c|}{\textbf{SDXL-Turbo}} & 
        \multicolumn{3}{c:}{\textbf{SD v2.1}} & 
        \multicolumn{3}{c}{\textbf{SDXL}} \\
        ~ & 1 & 2 & 3 & 1 & 2 & 3 &
        10 & 15 & 20 & 10 & 15 & 20 \\

        \hline
        Vanilla & 
        0.2961 & 0.3059 & 0.3034 & 
        0.2587 & 0.2693 & 0.2666 & 
        0.2876 & 0.2908 & 0.2932 & 
        0.2830 & \tbest{0.2889} & \tbest{0.2904} \\

        \hdashline

        blk-0 & 
        \sbest{0.2987} & \best{0.3088} & \tbest{0.3045} & 
        / & / & / & 
        0.2860$\downarrow$ & 0.2915 & \sbest{0.2946} & 
        / & / & / \\
        
        blk-1 & 
        \best{0.2993} & 0.3066 & 0.3019$\downarrow$ & 
        \best{0.2591} & 0.2685$\downarrow$ & 0.2666 & 
        \tbest{0.2881} & 0.2908 & \tbest{0.2943} & 
        \sbest{0.2863} & \best{0.2907} & \best{0.2924} \\
        
        blk-2 & 
        0.2966 & 0.3065 & 0.3037 & 
        0.2584$\downarrow$ & 0.2686$\downarrow$ & 0.2661$\downarrow$ & 
        0.2880 & 0.2896$\downarrow$ & 0.2935 & 
        0.2842 & 0.2886$\downarrow$ & 0.2889$\downarrow$ \\
        
        blk-3 & 
        0.2962 & 0.3055$\downarrow$ & 0.3035 & 
        \sbest{0.2589} & \best{0.2697} & \tbest{0.2668} & 
        0.2876 & 0.2909 & 0.2934 & 
        0.2839 & 0.2877$\downarrow$ & 0.2898$\downarrow$ \\
        
        blk-4 & 
        0.2965 & \sbest{0.3074} & \sbest{0.3052} & 
        \tbest{0.2588} & \sbest{0.2695} & \best{0.2671} & 
        \sbest{0.2894} & \tbest{0.2917} & 0.2939 & 
        \tbest{0.2851} & \tbest{0.2889} & 0.2892$\downarrow$ \\
        
        blk-5 & 
        \tbest{0.2981} & 0.3066 & \best{0.3062} & 
        0.2576$\downarrow$ & \tbest{0.2694} & \sbest{0.2670} & 
        \best{0.2902} & \best{0.2923} & \best{0.2950} & 
        \best{0.2868} & \sbest{0.2904} & \best{0.2910} \\
        
        blk-6 & 
        0.2962 & \tbest{0.3072} & 0.3025$\downarrow$ & 
        / & / & / & 
        0.2880 & \sbest{0.2919} & 0.2927$\downarrow$ & 
        / & / & / \\
        
        \hline
    \end{tabular}
    \label{tab-comparison_between_different_models_and_steps}
    \vspace{-10pt}
\end{table*}

\section{Related Work}

\subsection{U-Net Mechanisms in Diffusion Models}

Recently, there has been growing interest in the interpretability of diffusion models, especially the functionality of U-Net. 
For instance,~\cite{mei2024unetsbeliefpropagationefficient} 
proposes a hypothesis regarding the specific role of each layer within U-Net.
Researchers have also explored U-Net's mechanism from the frequency domain. 
FreeU~\cite{si2023freeufreelunchdiffusion} examines the component variation of different frequencies during the denoising process, pointing out that the U-Net backbone primarily contributes to denoising, while skip connections introduce high-frequency features into the decoder. 
Similarly,~\cite{yang2022diffusionprobabilisticmodelslim} found DMs are inclined to avoid generating high-frequency features, and learn to recover components of varying frequencies at different time steps. 

Orthogonal to the aforementioned approaches, we propose to monitor the importance variations of the Transformer blocks within the diffusion U-Net throughout the denoising process, which enables us to infer the underlying mechanisms of U-Net's components in image applications. 

\subsection{Training-free U-Net Capability Enhancement}

Enhancing U-Net's performance in image generation is another research focus. 
Unlike prior works~\cite{lee2023aligning, wallace2024diffusion}, which necessitate 
computationally intensive training processes, recent research has shifted focus towards leveraging the intrinsic mechanisms of the U-Net to enhance its capabilities without additional training or fine-tuning. 
For example, FreeU~\cite{si2023freeufreelunchdiffusion} 
effectively improves the sample quality by simply re-weighting the contributions from the skip connections and the backbone network. 
~\cite{liu2024understandingcrossselfattentionstable} achieved prompt-free real-image editing by replacing the self-attention maps without additional fine-tuning. 

Similar to FreeU~\cite{si2023freeufreelunchdiffusion}, we propose to re-weight the outputs of the Transformer blocks within the U-Net according to dynamic importance to enhance the U-Net capabilities in a training-free manner, which represents another ``free lunch" following FreeU. 
Differently, we employ a dynamic time-variant re-weighting schedule, instead of the static one in FreeU. 
Empirical results underscore the significance of considering the dynamic role evolution of attention blocks. 

\section{Method}

\subsection{Preliminary}
\label{section-method_subsection-preliminary}

Given a clean sample $\pmb{x}_0$ and a variance schedule $\{\overline{\alpha}_t\}_{t = 0}^T$, the deterministic reverse step of DDIM~\cite{song2022denoisingdiffusionimplicitmodels} is
\begin{equation}
    \pmb{x}_{t - 1}
    = 
    \sqrt{\overline{\alpha}_{t - 1}} \left(\dfrac{\pmb{x}_t - \sqrt{1 - \overline{\alpha}_t} \hat{\pmb{\epsilon}}_t}{\sqrt{\overline{\alpha}_t}}\right) + \sqrt{1 - \overline{\alpha}_{t - 1}} \hat{\pmb{\epsilon}}_t
    , 
\end{equation}
where $T$ is the amount of training steps of DMs, 
and $\hat{\pmb{\epsilon}}_t = \pmb{\epsilon}_\theta (\pmb{x}_t, t)$ represents the noise predicted by the U-Net parameterized by $\theta$ at time step $t$. 

The signal-to-noise ratio (SNR) of this step is defined as
\begin{equation}
    \label{eq-snr}
    \text{SNR} (\pmb{x}_t) 
    = 
    ||\pmb{x}_0||^2 / \text{Var} (\Delta\pmb{\epsilon}_t)
    = 
    ||\pmb{x}_0||^2 / \text{Var} (\pmb{\epsilon}_t - \hat{\pmb{\epsilon}}_t)
    , 
\end{equation}
where $||\pmb{x}_0||^2$ represents the \emph{power} of the true signal $\pmb{x}_0$, and $\Delta \pmb{\epsilon}_t = \pmb{\epsilon}_t - \hat{\pmb{\epsilon}}_t$ denotes the \emph{error} between the true noise $\pmb{\epsilon}_t$ and the predicted noise $\hat{\pmb{\epsilon}}_t$. 

The output of the $i$-th Transformer block is modeled as
\begin{equation}
    \pmb{y}_i 
    = 
    \pmb{f}_i (\pmb{x}_0) + \pmb{g}_i (\pmb{\epsilon}_t) + \pmb{n}_i
    , 
\end{equation}
where:

\begin{itemize}
    \item $\pmb{f}_i (\pmb{x}_0)$: The feature components related to the signal $\pmb{x}_0$. 

    \item $\pmb{g}_i (\pmb{\epsilon}_t)$: The feature components related to the noise $\pmb{\epsilon}_t$. 

    \item $\pmb{n}_i$: The intrinsic noise of the Transformer block. 
\end{itemize}

The following proposition provides an estimate for $\text{Var} (\hat{\pmb{\epsilon}}_t)$. 

\begin{proposition}
    \label{prop-reweighting}
    \emph{(Proof in Appendix)} 
    The variance of the error
    \begin{equation}
        \label{eq-the_variance_of_error}
        \begin{split}
            & \text{Var}(\Delta \hat{\pmb{\epsilon}}_t) 
            \approx \textstyle\sum_i A_i^2 (w_i - 1)^2 
            \text{Var} (\pmb{g}_i (\pmb{\epsilon}_t) )
            \\ 
            &~~~~ + 
            \textstyle\sum_i A_i^2 w_i^2 \text{Var}(\pmb{f}_i (\pmb{x}_0)) 
            + 
            \textstyle\sum_i A_i^2 w_i^2 \text{Var}(\pmb{n}_i)
            , 
        \end{split}
    \end{equation}
    where $A_i$ denotes the mapping transformation from the output of the $i$-th Transformer block to the final noise prediction. 
\end{proposition}

\subsection{Re-weighting the Outputs of Transformer Blocks} 
\label{sec-method_subsec-reweighting}

We propose to re-weight the output of the $i$-th $(i = 0, 1, \cdots)$ Transformer block by a scaling factor $w_i > 0$ before passing it to the subsequent ResNet block (see Fig.~\ref{fig-weighted_u_net}). 

Intuitively, applying a weight $w > 1.0$ amplifies the effect of the attention mechanism~\cite{vaswani2023attentionneed} within the Transformer block, 
whereas applying a weight $w < 1.0$ attenuates it. 
More rigorously, in accordance with Prop.~\ref{prop-reweighting}, we aim to reduce $\text{Var}(\Delta \hat{\pmb{\epsilon}}_t)$ via re-weighting, thus enhancing the SNR. 

However, we empirically observed that assigning arbitrary weights greater that 1.0 to any block does not always lead to performance enhancement (see Tab.~\ref{tab-comparison_between_different_models_and_steps}). 
In some cases, it can even yield worse performance compared to the vanilla weighting schedule, i.e., $w_i = 1.0$ for all blocks. 

We note that, this arises because the importance of blocks dynamically shifts throughout the denoising process (see Sec.~\ref{section-experiment_subsection-importance_ranking}). 
Statically assigning fixed weights to certain blocks may misweight the contributions of components in Eq.~\ref{eq-the_variance_of_error}, thus increasing $\text{Var}(\Delta \hat{\pmb{\epsilon}}_t)$, and consequently reducing the SNR. 
This led us to uncover and quantify the dynamic importance of each Transformer block thoughout the denoising process. 

\subsection{Importance Probe} 
\label{section-method_subsection-importance_probe}

A straightforward way to identify the importance of a block is to mask it out during inference. However, owing to the highly coupled functionality of the Transformer blocks within the U-Net, we cannot quantify block importance by this simple strategy.
In this paper, we propose \emph{Importance Probe} (IP), a novel technique to monitor the significance of each U-Net Transformer block. 
Specifically, we assign a non-negative weight to scale the output of each block at every inference step to measure its importance throughout the denoising process. 
In addition, each block is also associated with a weight threshold, which is dynamically adjusted during the probing process. 

To simplify, we restrict the weights and thresholds to real numbers in the range $[0, 1]$. 
If the weight of a block falls below its threshold, the attention computation in that block is skipped; otherwise, the attention is computed as usual, with the output of the Transformer block scaled by the block's weight. 
This strategy limits the capacity of the attention mechanism through a weight-threshold schedule. 

Then, our goal is to identify an optimal non-negative function
for each block with respect to the inference step, which reflects its importance. 
Specifically, a higher threshold indicates lower importance.
However, these thresholds cannot be directly obtained using standard optimization methods, as deciding whether to skip the attention computation in each block introduces non-differentiability into the optimization. 

To overcome this challenge, we employ a randomized heuristic search approach. 
We start by initializing all the block weights with uniformly sampled random values from the range $[0.99, 1.0]$, and all the block thresholds to $0.0$. 
Firstly, for the original U-Net, referred to as the ``\emph{teacher U-Net}'', the initial weights and thresholds are fixed for all blocks during the procedure. 
Secondly, for the copy of the teacher U-Net, named the ``\emph{student U-Net}'', the block weights and thresholds will be dynamically updated during the optimization process. 

In each iteration, for every inference step, we randomly perturb the best historical weights within a specified range using the \emph{Weight Bias Schedule} to obtain several new sets of weights. 
Each new weight set is evaluated using a criterion function (we implement with  L2 loss) to assess the performance of the student U-Net under those weights, reflecting the current state of the thresholds. 
Specifically, the new weight set is accepted when the error between the noise predicted by the student U-Net and the teacher U-Net falls within the maximum allowed tolerance; otherwise, it is rejected. 
The \emph{Threshold Update Schedule} adjusts the thresholds based on the above assessment. 
If at least one acceptable weight set is found, it indicates that the current threshold will likely have room for growth and can be increased. 
In contrast, the current threshold may be too high and should be reduced.

\paragraph{Weight Bias Schedule}
The magnitude of the perturbations is set to increase linearly along the inference progress, which aims to constrain the student U-Net to follow the denoising trajectory of the teacher U-Net in the early stages, while encouraging the student U-Net to explore finer image details in the later stages independently. 

To avoid the averaging effects of arbitrary random perturbations and achieve faster convergence, we stipulate that the \emph{energy} of the weight set should decrease during the importance probe process. 
Particularly,
the \emph{energy} of a weight set $\pmb{w} \in [0, 1]^m$ is defined as $E(\pmb{w}) = \sum_{i = 0}^{m - 1} w_i^2$, 

in which $w_0, \cdots, w_{m - 1}$ represents the weights for the $m$ target blocks ($m = 7/5$ for SD/SDXL U-Net) respectively. 
If multiple acceptable weight sets are found during an iteration, we retain the weight set with the highest fitness as the optimal solution. 
The \emph{fitness} of a weight set $\pmb{w}$ is defined as, 
\begin{equation}
    \text{fitness}(\pmb{w}) 
    = 
    E_0 / E(\pmb{w}) + 1/m \cdot \textstyle\sum\nolimits_{i = 0}^{m - 1} [w_i < q_i], 
    \label{eq-fitness}
\end{equation}
where $E_0$ stands for the initial energy of the system, $q_i$ denotes the current threshold for the $i$-th target block.
The term $[w_i < q_i]$ is under the Iverson bracket notation. 

\paragraph{Threshold Update Schedule}
To obtain a smoother threshold update, instead of performing a hard or soft update based on the performance of the student U-Net, we update  with conditional expectation. 
Refer to the Appendix for details. 

\subsection{Quantify Block Importance via the Voting Mechanism}

In task-specific scenarios, such as a text-to-image task with a 
fixed text prompt, IP can be employed to monitor the importance of the Transformer blocks at each step. 
Since the derived importance ranking may depend on the initial weight configuration, multiple runs with different weight initializations are conducted. 
The results are aggregated through a \emph{voting mechanism} to determine the final \emph{importance ranking}. 

For each run, the indices of the blocks are sorted according to their importance thresholds in descending order, resulting in a sequence $[idx_0, \cdots, idx_{m - 1}]$. 
It reflects the importance ranking, where blocks with higher indices are deemed more important. 
For this run, the $idx_i$-th block gains a score of $(i + 1)$. 
The final score for each block, named \emph{voting score}, is obtained by summing the scores across all runs, and blocks with higher cumulative scores are considered more important.

At inference step $t$, the \emph{importance score} of $i$-th block 
\begin{equation}
    is_i^{(t)}
    = 
    vs_i^{(t)} / (m \cdot r), 
\end{equation}
in which 
$r$ is the number of runs, $vs_i^{(t)}$ is the voting score.

\subsection{Adaptive Importance-based Re-weighting Schedule}

With the importance ranking, 
we designed an adaptive, importance-based re-weighting schedule to enhance the U-Net's capability in image generation tasks. 
For a specific text-to-image task, we first evaluate the importance of each block using several runs of IP. 
At each step $t$, we quantify to obtain the importance score for each block $[is_0^{(t)}, \cdots, is_{m - 1}^{(t)}]$. 

Subsequently, we select and fix a weight range $[low, high]$, 
and the \emph{weight} of $i$-th block at step $t$ is assigned as
\begin{equation}
    w_{i}^{(t)}
    = 
    \begin{cases}
        is_i^{(t)} \cdot (high - low) + low & low \neq high \\
        high & low = high
    \end{cases}. 
\end{equation}

We perform the denoising process as usual, in which we scale the output of the $i$-th block by $w_i^{(t)}$ at step $t$. 
The entire process described above is training-free. 

\subsection{Empirical Re-weighting Strategy}
\label{sec-method_subsection-empirical_reweighting_strategy}

By applying the importance-based re-weighting schedule, we can assign greater weights to dominant blocks at each step, thereby increasing the SNR. Empirically: 

\begin{itemize}
    \item It is more likely to assign greater weights to bottleneck blocks, as they encode high-level features, and serve as the nexus between the encoder and decoder.

    \item In the early denoising, it is more likely to assign greater weights to mid-low-resolution blocks, emphasizing terms with smaller $\text{Var} (\pmb{f}_i)$. 

    \item In the later denoising, it is more likely to assign greater weights to high-resolution blocks, emphasizing terms with smaller $\text{Var} (\pmb{g}_i)$. 

    \item Throughout denoising, it is more likely to assign smaller weights to blocks with larger intrinsic noise, suppressing terms with larger $\text{Var} (\pmb{n}_i)$.
\end{itemize}

\subsection{Dynamic Attention Pruning Tests}

Due to the absence of prior work on dynamic importance ranking as a reference, we further validate the derived importance ranking through dynamic attention pruning tests. 
These experiments utilize the importance ranking to design pruning strategies tailored to specific tasks. 
Specifically, the dynamic pruning strategies involve skipping the one or two least important blocks at each step. 
By enumerating all possible combinations, we generate a series of pruning strategies.

We dynamically prune the student U-Net according to each pruning strategy, and fine-tune the student U-Net under the supervision of the teacher U-Net. 
During this process, we freeze the parameters of all U-Net blocks except for the Transformer blocks. 
After fine-tuning, we compare the sampling results of the temporally pruned student U-Net with those of the complete teacher U-Net. 

\section{Experiment}

\subsection{Re-weighting Schedule across Various Models}
\label{sec-experiment_subsec-reweighting_schedule}

We compared the effects of a static weighting schedule, where each block is assigned a weight of 1.1 respectively, across different inference steps with SD-Turbo~\cite{sauer2024fasthighresolutionimagesynthesis}, SDXL-Turbo~\cite{sauer2023adversarialdiffusiondistillation}, SD~\cite{rombach2022highresolutionimagesynthesislatent} and SDXL~\cite{podell2023sdxlimprovinglatentdiffusion}. 
The SD/SDXL family generates images at a resolution of 512$\times$512/1024$\times$1024. 

Empirical results are presented in Tab.~\ref{tab-comparison_between_different_models_and_steps}, in which samples evaluated with the Human Preference Score v2 (HPS v2)~\cite{wu2023humanpreferencescorev2} (the higher, the better).
The results demonstrate that, across all configurations (each column), there exists at least one weighted schedule that outperforms the vanilla one in terms of aesthetics. 
For each weighting schedule (each row), most instances yield higher aesthetic scores than the vanilla one, while the scores get lower in some cases. 
On the one hand, this highlights the robustness of our method in enhancing sample aesthetics across different models, inference steps, and sample resolutions. 
On the other hand, it illustrates that, simply re-weighting arbitrary blocks is not sufficient to guarantee an improved SNR during the denoising process. 

Additionally, it can also be observed that, in experiments with all models except SD v2.1, instances occur where the performance with re-weighting at the second-highest inference step surpasses the performance without re-weighting at the highest inference step. 
Qualitative results shown in Fig.~\ref{fig-head_figure} (a) indicate that, our method not only enhances sample quality, but also improves sampling efficiency, which is attributed to the increased SNR during the denoising process. 

\subsection{Importance Ranking}
\label{section-experiment_subsection-importance_ranking}

We select the text-to-image generation with a fixed text prompt ``\emph{Some cut up fruit is sitting in a blender. }" as our task. 
We sample with 2-step inference SD-Turbo and SDXL-Turbo respectively, and derive the dynamic importance using IP and the voting mechanism. 
Results are listed in Tab.~\ref{tab-dynamic_importance_ranking_fruit}. 

It demonstrates that the blocks exhibit dynamic importance shifts throughout the denoising process, indicating that their roles evolve over time. 
Moreover, we observe that, the importance of symmetrically positioned blocks often shows dramatic disparities, suggesting that the significance of architecturally symmetric blocks is not fully aligned. 
Specifically, within symmetrically positioned pairs, the blocks belonging to the decoder ($idx = 4, 5, 6$) tend to be more important, while the mid-block ($idx = 3$) consistently maintains high importance. 

During the inference process, bottleneck blocks consistently maintain a high level of importance. 
In the early denoising, mid-low resolution blocks exhibit greater significance, while in the later stages, the importance of high-resolution blocks relatively increases.
This experimental result aligns with the re-weighting strategy outlined in Sec.~\ref{sec-method_subsection-empirical_reweighting_strategy}. 

\begin{table}[h]
    \centering
    \caption{
        Dynamic importance ranking of 2-step SD-Turbo/SDXL-Turbo U-Net, arranged in non-descending order. 
    }
    \begin{tabular}{c | c : c}
        \hline
        \multirow{2}{*}{\textbf{Step}} & 
        \multicolumn{2}{c}{\textbf{Importance Ranking}} \\

        ~ & SD-Turbo & SDXL-Turbo \\
        \hline

        0 & 0 1 2 4 6 5 3 & 1 2 5 3 4 \\ 
        1 & 1 0 5 4 2 6 3 & 1 2 5 4 3 \\ 

        \hline
    \end{tabular}
    \label{tab-dynamic_importance_ranking_fruit}
    \vspace{-10pt}
\end{table}

\subsection{Dynamic Attention Pruning Tests}

We conduct dynamic attention pruning tests to validate the derived importance ranking. 
we assess how removing an equal number of blocks under different skipping strategies impacts the model's performance. 
Specifically, we benchmark our method against various static, dynamic, symmetric, and unnecessary-symmetric skipping strategies. 

\begin{figure}[h]
    \centering
    \includegraphics[width=0.98\linewidth]{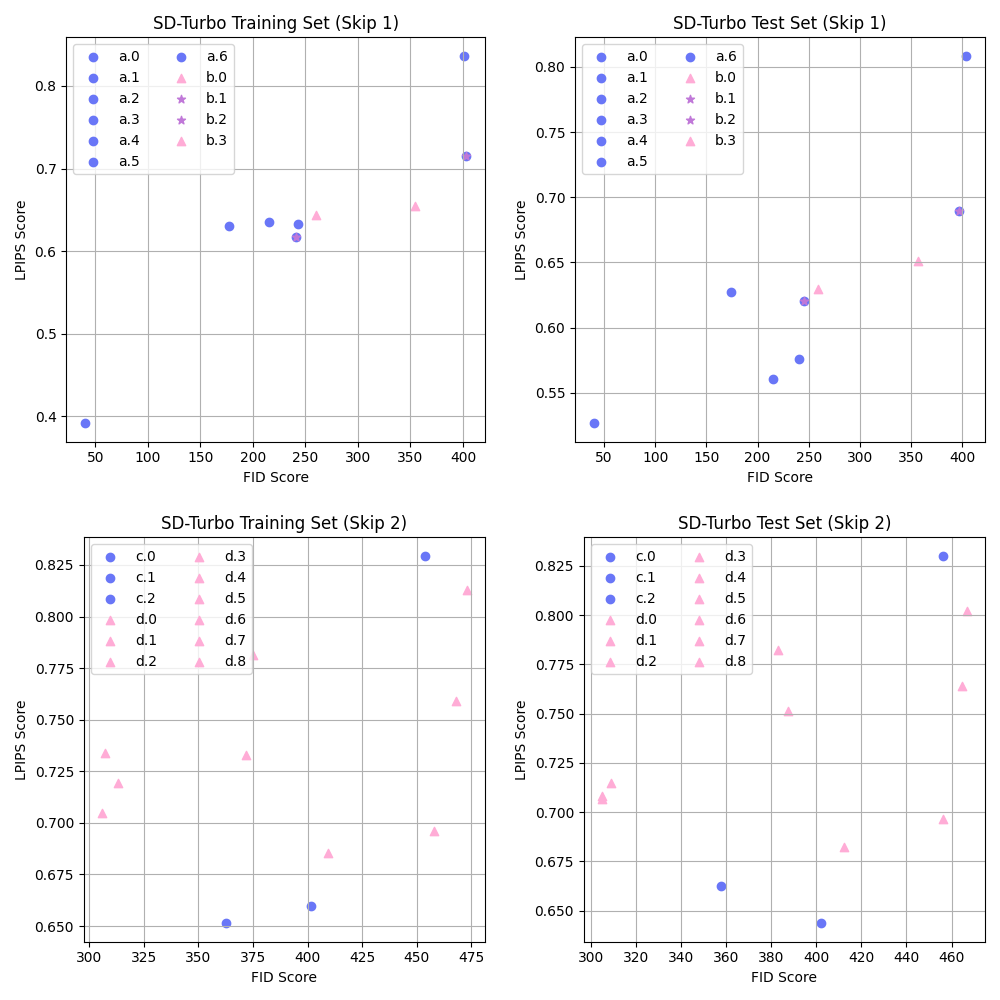}
    \caption{
        Scatter plot of FID and LPIPS under different skipping strategies (the further lower-left, the better). 
        Baseline strategies are represented by blue circles, unique points from our strategies are shown as pink triangles, while points overlapping with baseline points are marked with purple stars.
    }
    \label{fig-scatter_fid_lpips_fruit}
    \vspace{-10pt}
\end{figure}

Results are plotted in Fig.~\ref{fig-scatter_fid_lpips_fruit}. 
Our skipping strategies achieve better performance than baseline strategies, especially in cases where two blocks are skipped per inference step, which validate the correctness of the obtained importance ranking. 

Qualitative results shown in Fig.~\ref{fig-head_figure} (d) indicate that, 
our dynamic approaches achieve better fidelity in pruned sampling. 

\subsection{Enhanced Image Synthesis}

We benchmark our method with Human Preference Dataset v2~\cite{wu2023humanpreferencescorev2}, from which we randomly sampled 200 prompts in each category. 
For each prompt, we generated 1 sample in 2 inference steps with SD-Turbo and SDXL-Turbo respectively. 
We evaluate the samples using HPS v2. 
We select 42 and 21 as the seeds for the training and test sets respectively. 

\begin{figure}[h]
    \centering
    \includegraphics[width=0.98\linewidth]{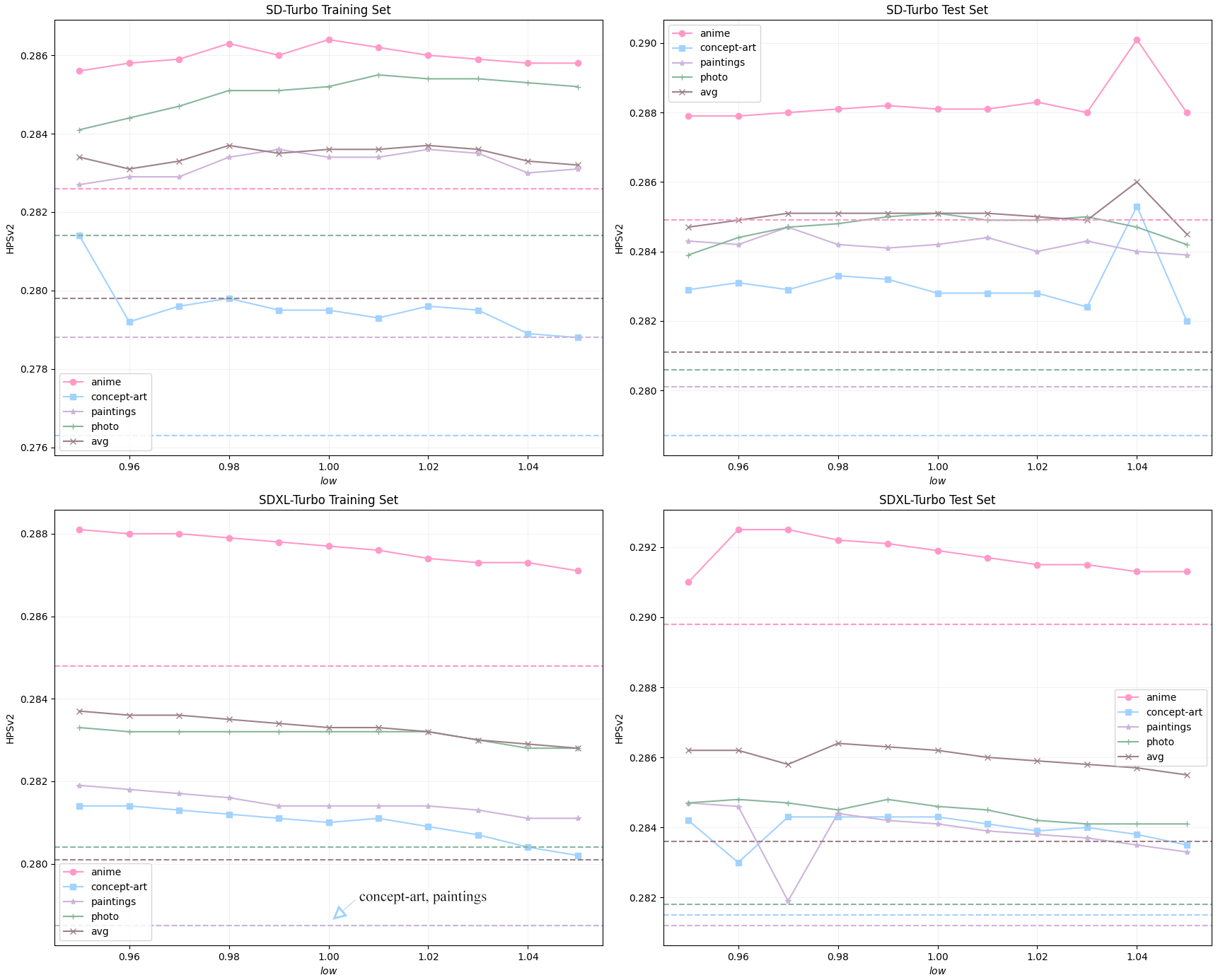}
    \caption{
        Line chart showing the effect of re-weighting on SD-Turbo and SDXL-Turbo with fixed $high = 1.1$ as $low$ varies. 
        Lines of the same color represent the same category, where dashed lines indicate the vanilla schedule, and solid lines represent our re-weighting schedule. 
    }
    \label{fig-hpd_sdturbo_sdxlturbo_fix_r}
    \vspace{-5pt}
\end{figure}

Firstly, we fix $high = 1.1$ for the weight range, and investigate the impact of varying $low \in [0.95, 1.05]$. 
The variation of aesthetic scores with respect to $low$ is plotted in Fig.~\ref{fig-hpd_sdturbo_sdxlturbo_fix_r}. 
The results indicate that our 
re-weighting schedule consistently outperforms the vanilla one in both the training and test set, demonstrating the robustness and generalization of our method across different categories of prompts. 

Subsequently, we fix the optimal $low$ for each model, specifically, $low = 0.98$ and $1.02$ for SD-Turbo, and $low = 0.95$ for SDXL-Turbo. 
We explore the effects of varying $high \in \{1.11, 1.15, 1.2\}$. 
Quantitative results are presented in Tab.~\ref{tab-hpd_sdturbo_selected} and \ref{tab-hpd_sdxlturbo_selected}. 
It shows that re-weighting schedules with $low$ slightly below 1.0 generally yield better performance. 

The quantitative results also reveal that, excessively high values of $high$ lead to performance degradation. 
To illustrate this, we present qualitative results in Fig.~\ref{fig-head_figure} (b). 
It can be observed that our method significantly enhances aesthetics when the weight range is appropriately chosen. 
However, when the value of $high$ is too large (e.g. $high = 1.2$), the samples exhibit color oversaturation, blurring, and artifacts, leading to a decrease in aesthetic quality.

\subsection{Ablation Study}

Table~\ref{tab-comparison_between_different_models_and_steps} already demonstrates that, arbitrary re-weighting does not guarantee performance improvement. 

We conducted another 
alabtion study by inverting the importance scores 
with the optimal weight ranges.
Specifically, we compute the \emph{inverted} importance scores as 
\begin{equation}
    \overline{is}_i^{(t)}
    = 
    (m \cdot r - vs_i^{(t)}) / (m \cdot r), 
\end{equation}
and use them to sample the weights. 

Results are listed in Tab.~\ref{tab-sd_turbo_best_ablation} and \ref{tab-sdxl_turbo_best_ablation}. 
It illustrate that, in the majority of cases, the performance declined after inverting the importance scores, but still remained higher than that of the vanilla schedule. 
This validates the necessity of accounting for the importance ranking in enhancing the U-Net capability. 

\subsection{Comparison with FreeU}

Both our method and FreeU~\cite{si2023freeufreelunchdiffusion} enhance the U-Net capacity through re-weighting. 
However, FreeU employs a static re-weighting schedule that is agnostic to the importance of components. 
This approach, though improving sample aesthetics, may struggle with preserving the identity (see Fig.~\ref{fig-head_figure} (c)).

We hypothesize that this discrepancy arises, because our method accounts for task-specific importance, providing a more fine-grained and moderate enhancement. 
In contrast, FreeU's simultaneous scaling of multiple components introduces a more aggressive impact. 

\section{Conclusion}

In this study, we assessed the dynamic importance of Transformer blocks within the diffusion U-Net with Importance Probe. 
By temporally scaling the output of Transformer blocks based on an adaptive importance-based re-weighting schedule, we achieved capability enhancement for the U-Net in image synthesis scenarios. 
These findings demonstrate the potential of incorporating dynamic mechanisms to improve the performance of diffusion models across various applications. 

\begin{minipage}{1.035\columnwidth}
    \centering
    
    \captionof{table}{
        Enhanced SD-Turbo (selected). 
    }
    \label{tab-hpd_sdturbo_selected}
    \hspace{-3em}
    \resizebox{0.98\columnwidth}{!}{
        \centering
        \begin{tabular}{c | c c c c c c c c | c c}
            \hline
            \multirow{2}{*}{\textbf{Weighting}} &
            \multicolumn{2}{c}{\textbf{anime}} & 
            \multicolumn{2}{c}{\textbf{concept-art}} & 
            \multicolumn{2}{c}{\textbf{paintings}} & 
            \multicolumn{2}{c|}{\textbf{photo}} &
            \multicolumn{2}{c}{\textbf{avg}} \\
    
            ~ & 
            train & test & train & test &
            train & test & train & test & 
            train & test \\
    
            \hline

            Vanilla & 
            0.2826 & 0.2849 &
            0.2763 & 0.2787 &
            0.2788 & 0.2801 &
            0.2814 & 0.2806 & 
            0.2798 & 0.2811 \\

            \hdashline

            $[0.98, 1.1]$ & 
             \best{0.2863} & \sbest{0.2881} & 
             \best{0.2798} & \best{0.2833} &
             \sbest{0.2834} & \best{0.2842} &
             \sbest{0.2851} & \sbest{0.2848} & 
             \best{0.2837} & \best{0.2851} \\

             $[1.02, 1.1]$ & 
             \sbest{0.2860} & \best{0.2883} & 
             \sbest{0.2796} & \sbest{0.2828} &
             \best{0.2836} & \sbest{0.2840} &
             \best{0.2854} & \best{0.2849} & 
             \best{0.2837} & \sbest{0.2850} \\

            \hdashline

            $[0.98, 1.15]$ & 
             \tbest{0.2859} & \tbest{0.2875} & 
             \tbest{0.2793} & \tbest{0.2822} &
             \tbest{0.2833} & \tbest{0.2839} &
             \tbest{0.2845} & \tbest{0.2840} & 
             \sbest{0.2833} & \tbest{0.2844} \\

             $[1.02, 1.15]$ & 
             0.2853 & 0.2869 & 
             0.2783 & 0.2809 &
             0.2822 & 0.2832 &
             0.2843 & 0.2830 & 
             \tbest{0.2825} & 0.2835 \\

             \hdashline

             $[0.98, 1.2]$ & 
             0.2852 & 0.2860 & 
             0.2776 & 0.2802 &
             0.2815 & 0.2823 &
             0.2828 & 0.2815 & 
             0.2818 & 0.2825 \\

             $[1.02, 1.2]$ & 
             0.2833 & 0.2837 & 
             0.2758 & 0.2786 &
             0.2802 & 0.2809 &
             0.2815 & 0.2794 & 
             0.2802 & 0.2806 \\

            \hline
        \end{tabular}
    }

    \captionof{table}{
        Enhanced SDXL-Turbo (selected). 
    }
    \label{tab-hpd_sdxlturbo_selected}
    \hspace{-3em}
    \resizebox{0.98\columnwidth}{!}{
        \centering
        \begin{tabular}{c | c c c c c c c c | c c}
            \hline
            \multirow{2}{*}{\textbf{Weighting}} &
            \multicolumn{2}{c}{\textbf{anime}} & 
            \multicolumn{2}{c}{\textbf{concept-art}} & 
            \multicolumn{2}{c}{\textbf{paintings}} & 
            \multicolumn{2}{c|}{\textbf{photo}} &
            \multicolumn{2}{c}{\textbf{avg}} \\
    
            ~ & 
            train & test & train & test &
            train & test & train & test & 
            train & test \\
    
            \hline

            Vanilla & 
            \tbest{0.2848} & 0.2898 &
            \tbest{0.2785} & 0.2815 &
            \tbest{0.2785} & \tbest{0.2812} &
            \tbest{0.2804} & \tbest{0.2818} & 
            \tbest{0.2801} & 0.2836 \\

            \hdashline

            $[0.95, 1.1]$ & 
             \best{0.2881} & \tbest{0.2910} & 
             \best{0.2814} & \sbest{0.2842} &
             \best{0.2819} & \best{0.2847} &
             \best{0.2833} & \sbest{0.2847} & 
             \best{0.2837} & \sbest{0.2862} \\

            $[0.95, 1.15]$ & 
             \best{0.2881} & \best{0.2923} & 
             \best{0.2814} & \best{0.2844} &
             \sbest{0.2818} & \sbest{0.2844} &
             \best{0.2833} & \best{0.2848} & 
             \best{0.2837} & \best{0.2865} \\

             $[0.95, 1.2]$ & 
             \sbest{0.2878} & \sbest{0.2916} & 
             \sbest{0.2810} & \tbest{0.2838} &
             \best{0.2816} & \best{0.2839} &
             \sbest{0.2826} & \best{0.2842} & 
             \sbest{0.2832} & \tbest{0.2859} \\
        
            \hline
        \end{tabular}
    }
\end{minipage}

\begin{minipage}{1.035\columnwidth}
    \centering
    \vspace{5pt}
    \captionof{table}{
        SD-Turbo Ablation. 
    }
    \label{tab-sd_turbo_best_ablation}
    \hspace{-3em}
    \resizebox{0.98\columnwidth}{!}{
        \centering
        \begin{tabular}{c | c c c c c c c c | c c}
            \hline
            \multirow{2}{*}{\textbf{Weighting}} &
            \multicolumn{2}{c}{\textbf{anime}} & 
            \multicolumn{2}{c}{\textbf{concept-art}} & 
            \multicolumn{2}{c}{\textbf{paintings}} & 
            \multicolumn{2}{c|}{\textbf{photo}} &
            \multicolumn{2}{c}{\textbf{avg}} \\
    
            ~ & 
            train & test & train & test &
            train & test & train & test & 
            train & test \\
    
            \hline

            Vanilla & 
            0.2826 & 0.2849 &
            0.2763 & 0.2787 &
            0.2788 & \sbest{0.2801} &
            0.2814 & 0.2806 & 
            0.2798 & \sbest{0.2811} \\

            \hdashline

            $[0.98, 1.1]^{\text{rev}}$ & 
            \sbest{0.2857} & \best{0.2885} & 
            \sbest{0.2789} & \sbest{0.2826} &
            \sbest{0.2831} & \best{0.2842} &
            \best{0.2857} & \best{0.2853} & 
            \sbest{0.2834} & \best{0.2851} \\

            $[0.98, 1.1]$ & 
            \best{0.2863} & \sbest{0.2881} & 
            \best{0.2798} & \best{0.2833} &
            \best{0.2834} & \best{0.2842} &
            \sbest{0.2851} & \sbest{0.2848} & 
            \best{0.2837} & \best{0.2851} \\

            \hline
        \end{tabular}
    }

    \captionof{table}{
        SDXL-Turbo Ablation. 
    }
    \label{tab-sdxl_turbo_best_ablation}
    \hspace{-3em}
    \resizebox{0.98\columnwidth}{!}{
        \centering
        \begin{tabular}{c | c c c c c c c c | c c}
            \hline
            \multirow{2}{*}{\textbf{Weighting}} &
            \multicolumn{2}{c}{\textbf{anime}} & 
            \multicolumn{2}{c}{\textbf{concept-art}} & 
            \multicolumn{2}{c}{\textbf{paintings}} & 
            \multicolumn{2}{c|}{\textbf{photo}} &
            \multicolumn{2}{c}{\textbf{avg}} \\
    
            ~ & 
            train & test & train & test &
            train & test & train & test & 
            train & test \\
    
            \hline

            Vanilla & 
            0.2848 & 0.2898 &
            0.2785 & 0.2815 &
            0.2785 & 0.2812 &
            0.2804 & 0.2818 & 
            0.2801 & 0.2836 \\

            \hdashline

            $[0.95, 1.15]^{\text{rev}}$ & 
            \sbest{0.2861} & \sbest{0.2904} & 
            \sbest{0.2791} & \sbest{0.2831} &
            \sbest{0.2802} & \sbest{0.2830} &
            \sbest{0.2820} & \sbest{0.2837} & 
            \sbest{0.2818} & \sbest{0.2851} \\

            $[0.95, 1.15]$ & 
            \best{0.2881} & \best{0.2923} & 
            \best{0.2814} & \best{0.2844} &
            \best{0.2818} & \best{0.2844} &
            \best{0.2833} & \best{0.2848} & 
            \best{0.2837} & \best{0.2865} \\

            \hline
        \end{tabular}
    }
\end{minipage}


\section*{Acknowledgment}
This work was partially supported by the National Key R\&F Program of China (2024YFB3908500, 2024YFB
3908502, 2024YFB3908505), the DEGP Innovation Team (2022KCXTD025), and the Shenzhen University Teaching Reform Key Program (JG2024018).

\bibliographystyle{IEEEbib}
\bibliography{icme2025references}



\end{document}